%% file: arxiv.tex
\documentclass[10pt,twocolumn,letterpaper]{article}

\usepackage{iccv}

\usepackage{algorithm}
\usepackage[noend]{algpseudocode}
\usepackage{times}
\usepackage{epsfig}
\usepackage{graphicx}
\usepackage{bm}
\usepackage{amsmath,array}
\usepackage{amssymb}
\usepackage{dsfont}
\usepackage{comment}
\usepackage{bbm}
\usepackage{bbold}
\usepackage{adjustbox}
\usepackage{booktabs,subcaption,amsfonts,dcolumn}
\usepackage{xcolor,colortbl}

\usepackage{authblk}

\usepackage[pagebackref=true,breaklinks=true,letterpaper=true,colorlinks,bookmarks=false]{hyperref}

\DeclareMathOperator*{\argmax}{arg\,max}

\newcommand*\samethanks[1][\value{footnote}]{\footnotemark[#1]}
\definecolor{green_table}{rgb}{0.95, 1.0, 0.95}
\definecolor{red_table}{rgb}{1.0, 0.90, 0.90}

\iccvfinalcopy 

\def\iccvPaperID{1444} 

\ificcvfinal\fi

\begin{document}

\title{Conditional Variational Capsule Network for Open Set Recognition}


\author[2,1]{Yunrui Guo\thanks{Indicates equal contributions.}}
\author[1]{Guglielmo Camporese\samethanks}
\author[2]{Wenjing Yang}
\author[1]{Alessandro Sperduti}
\author[1]{Lamberto Ballan}
\affil[1]{Department of Mathematics ``Tullio Levi-Civita'', University of Padova, Italy}
\affil[2]{National University of Defense Technology, China\vspace*{-5pt}}


\maketitle

\input{./sections/s0_abstract_iccv.tex}

\input{./sections/s1_introduction_iccv.tex}

\input{./sections/s2_related_work_iccv.tex}

\input{./sections/s3_preliminaries_iccv.tex}

\input{./sections/s4_proposed_method_iccv.tex}

\input{./sections/s5_experimental_results_iccv.tex}
\input{./sections/s6_conclusion_iccv.tex}


\paragraph*{Acknowledgements.}
This work was supported in part by the PRIN-17 PREVUE project, from the Italian MUR (CUP: E94I19000650001).
YG was supported by a CSC fellowship.
We also acknowledge the HPC resources of UniPD -- DM and CAPRI clusters -- and the support of NVIDIA for their donation of GPUs used in this research.
Finally, we would like to thank the anonymous
reviewers for their valuable comments and suggestions.

{\small
\bibliographystyle{ieee_fullname}
\bibliography{openset}
}

\end{document}

%% file: sections/s0_abstract_iccv.tex
\begin{abstract}
In open set recognition, a classifier has to detect unknown classes that are not known at training time. In order to recognize new categories, the classifier has to project the input samples of known classes in very compact and separated regions of the features space for discriminating samples of unknown classes.
Recently proposed Capsule Networks have shown to outperform alternatives in many fields, particularly in image recognition, however they have not been fully applied yet to open-set recognition.
In capsule networks, scalar neurons are replaced by capsule vectors or matrices, whose entries represent different properties of objects. In our proposal, during training, capsules features of the same known class are encouraged to match a pre-defined gaussian, one for each class. To this end, we use the variational autoencoder framework, with a set of gaussian priors as the approximation for the posterior distribution. In this way, we are able to control the compactness of the features of the same class around the center of the gaussians, thus controlling the ability of the classifier in detecting samples from unknown classes. We conducted several experiments and ablation of our model, obtaining state of the art results on different datasets in the open set recognition and unknown detection tasks.
\end{abstract}

%% file: sections/s1_introduction_iccv.tex
\section{Introduction}

\begin{figure}[ht]
    \centering
    \includegraphics[width=0.95\columnwidth]{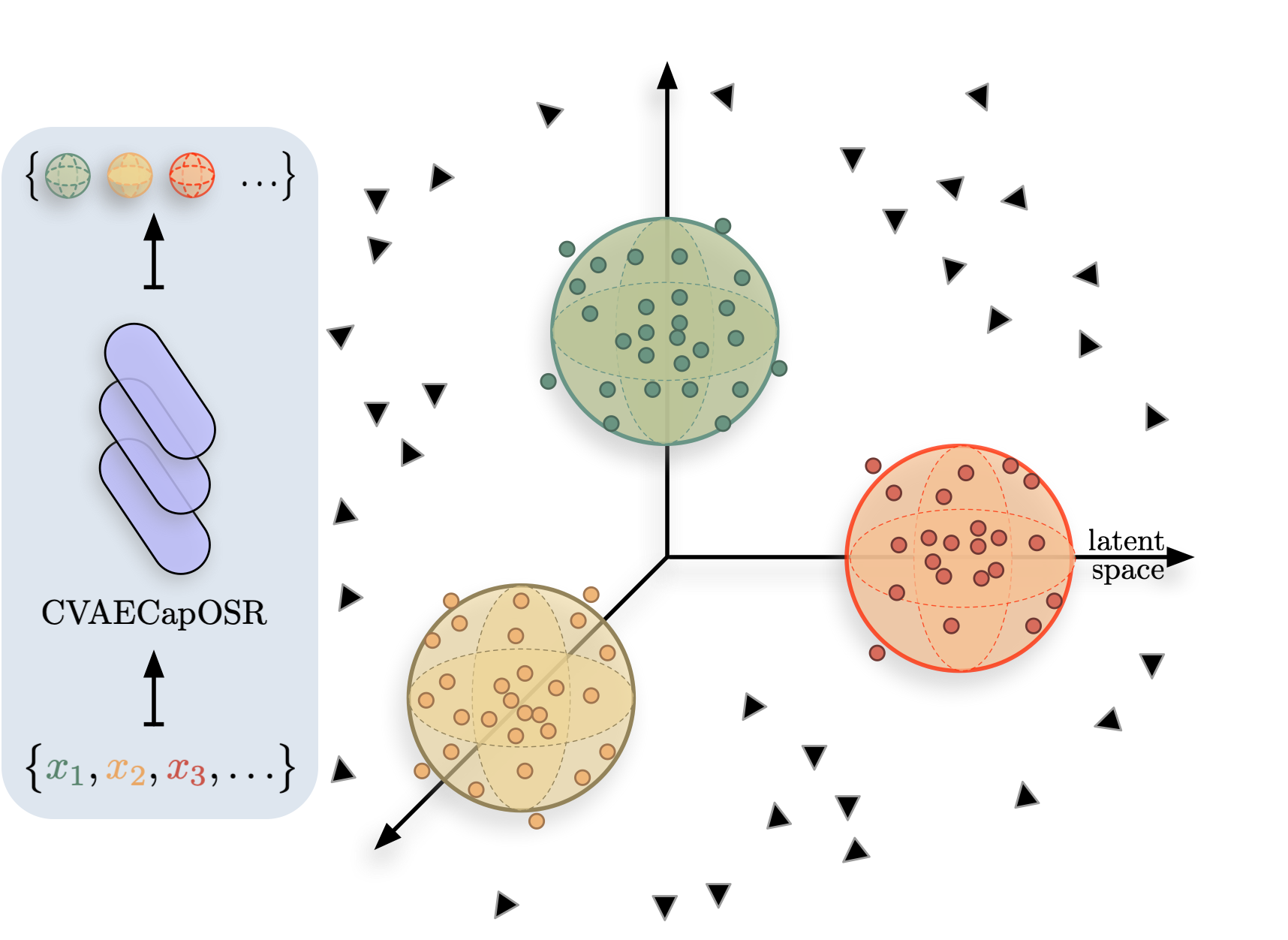} \\
    \caption{CVAECapOSR Model. Input samples are fed into the Capsule Network that produces distributions over the latent space. Each class has its own prior gaussian distribution in the feature space, that in the figure are represented as spheres. After training, the known samples (represented as small points) are clustered around the class target gaussians. The samples belonging to unknown classes are represented as black triangles, far from the target distributions.\label{fig:VAECapOSR}}
\end{figure}

Over the past decade, deep learning has become the dominant approach in many computer vision problems, achieving spectacular results on many visual recognition tasks \cite{imagenet,vgg16,resnet,deepface}.
However, most of these results have been obtained in a closed set scenario, where a critical assumption is that all samples should belong to at least one labeled category.
When observing a sample from an unknown category, closed-set approaches are forced to choose a class label from one of the known classes, thus limiting their applicability in dynamic and ever-changing scenarios.

To overcome such a limitation, open set recognition has been introduced to enable a classification system to identify all of known categories, while simultaneously detecting unknown test samples \cite{tosr,towr}.
In the open set scenario, samples included/excluded in label space are referred to as knowns/unknowns.
Therefore, open set classifiers need to use incomplete knowledge learned from a finite set of accessible categories to devise effective representations able to separate knowns from unknowns.
Early works have identified this issue, thus proposing methods employing different thresholding strategies for rejection of unknowns \cite{tosr, towr}.

Deep neural networks, despite demonstrating strong capabilities in learning discriminative representations in closed scenarios, show accuracy degradation within open set settings \cite{topn}.
As a naive strategy, modeling a threshold for Softmax outputs has been demonstrated to be a sub-optimal solution for deep neural networks to identify unknowns. Thus the Extreme Value Theory was introduced to better adapt these discriminative models, fully based on supervised learning, for open-set settings. The underpinning idea is to calibrate Softmax scores so to estimate the probability of unknowns \cite{crosr,topn}. In addition to deep discriminative models, deep generative models focusing on learning efficient latent feature representations by unsupervised leaning, have been widely utilized in open-set recognition tasks, and have gained  successes one after the other \cite{Neal_2018_ECCV,c2ae,Perera_2020_CVPR,cgdl}.
In particular, The Variational Auto-Encoder (VAE) is a typical probabilistic generative model ideal for detecting unknowns, due to its ability in learning low-dimensional representations in latent space not only supporting input reconstruction but also approximating a specified prior distribution. On the other hand, the VAE-based models may be not sufficiently effective for identifying known categories as all feature representations only follow one distribution. To this end, we employ a Conditional VAE (CVAE) that uses multiple prior distributions for modeling the known classes, and indirectly the unknown counterpart. Furthermore, we propose to represent the input samples with probabilistic capsules, given their already proved representation power capability~\cite{Rajasegaran_2019_CVPR, sabour2017dynamic}.

Capsule Networks (CapsNet) \cite{sabour2017dynamic} were proposed as an alternative to Convolutional Neural Networks (CNNs).
Unlike CNNs’ scalar neurons, capsules ensemble a group of neurons to accept and output vectors. The vector of an activated capsule represents the various properties of a particular object, such as position, size, orientation, texture, etc. In essence, CapsNet can be viewed as an encoder encoding objects by distributed representations, which is exponentially more efficient than encoding them by activating a single neuron in a high-dimensional space. Besides, CapsNet has been successfully used to detect fake images and videos in a task setting similar to open set recognition \cite{fake}.
This motivated us to design a novel capsule network architecture in combination with CVAE for the open set recognition problem, dubbed CVAECapOSR, that is depicted in Figure~\ref{fig:VAECapOSR}. 


The contributions of this paper are three-fold:
\emph{i)}~We present a novel open set recognition framework based on CapsNet and show its advantages for learning an efficient representation for known classes.
\emph{ii)}~We integrate CapsNet and conditional VAEs. In contrast to general VAEs that encourage the latent representation to approximate a single prior distribution, our model exploits multiple priors (i.e. one for each class), and it forces the latent representation to follow the gaussian prior selected by the class of the input sample.
\emph{iii)}~We conduct extensive experiments on all the standard datasets used for open set recognition, obtaining very competitive results, that in several cases outperform previous state of the art methods by a large margin.

%% file: sections/s2_related_work_iccv.tex
\section{Related Work}
The open set recognition problem was introduced by~\cite{tosr} and was initially formalized as a constrained minimization problem based on Support Vector Machines (SVMs), whereas subsequent works focused on other more traditional approaches, such as Extreme Value Theory~\cite{probinclu, probmod}, sparse representation~\cite{srosr}, and Nearest Neighbors~\cite{nearest}.

Following the success achieved by deep learning in many computer vision tasks, deep networks were first introduced for open set recognition in~\cite{topn}, in which it is proposed an Openmax function by calibrating the Softmax probability of each class with a Weibull distribution model.
Subsequently, \cite{gopenmax} extended Openmax to G-Openmax by introducing a generative adversarial network in which the generator produces synthetic samples of novel categories and the discriminator learns the explicit representation for unknown classes.
A similar strategy has been adopted in~\cite{Neal_2018_ECCV}, that presented a data augmentation technique based on generative adversarial networks, referred as counterfactual images generation. 
More recently, Yoshihashi \etal~\cite{crosr} analyzed and demonstrated the usefulness of training deep networks jointly for classification and reconstruction in the open set scenario. Specifically, the authors proposed to separate the knowns from the unknowns using the representations produced by the unsupervised training, while maintaining the discrimination capability of the model using the representations computed via the supervised learning process.

C2AE~\cite{c2ae} introduced an architecture for open set recognition and unknown detection based on class conditioned VAEs by modelling the reconstruction error of the model based on the Extreme Value Theory. 
Sun \etal~\cite{cgdl} have recently argued that one disadvantage of VAE-based architectures for open set recognition is the inadequate discriminative ability on instances of known classes. Therefore, the authors employed a conditional Gaussian distribution VAE model for learning conditional distributions of known classes and rejecting unknowns.
A different approach is presented in~\cite{hybrid}, where normalizing flows are employed for density estimation of known samples. Specifically, the authors proposed an architecture that uses a CNN encoder and an invertible neural network that jointly learns the density of the input. However, a potential issue not discussed in the paper is that the CNN encoder has no bijective property, that is crucial to employ the change-of-variables formula for density evaluation.
Additionally, \cite{rpl} introduced the concept of reciprocal points in prototype leaning to manage the open space. Although this work shows excellent performances in rejecting unknowns coming from a different dataset with respect to the known samples, the unknown detection capability degradates when the source of unknown samples is the same of the known counterpart.


%% file: sections/s3_preliminaries_iccv.tex
\section{Preliminaries}

\subsection{The Open Set Recognition Problem}
In the open set recognition problem the model has to classify test samples that can belong to classes not seen during training. Given a classification dataset \mbox{$D = \{(\bm{x}_1, y_1), \dots, (\bm{x}_n, y_n) \}$} such that \mbox{$\bm{x}_i \in \mathcal{X}$} is an input sample, $y_i \in \{1, \dots, K\}$ is the corresponding category label, the open set problem consists in the classification of test samples among $K + U$ classes, where the $U$ are the number of unknown classes. In the literature, the dataset used for training is called \textit{closed} dataset, meanwhile the one used during evaluation, that contains samples from unseen classes, is called \textit{open} dataset. In order to quantify the openness of a dataset during the evaluation, following \cite{tosr,Neal_2018_ECCV} we consider the \textit{openness} measure as $O = 1 - \sqrt{\frac{K}{M}}$ where $K$ and $M = K + U$ are  the number of classes observed during training and test, respectively. 

\subsection{Conditional VAE Formulation}
The Conditional Variational Eutoencoder (CVAE) directly derives from the VAE model \cite{vae} and its objective is based on the estimation of the conditioned density $p(x|y)$ of the data $x$ given the label $y$. It is one of the most powerful probabilistic generative models for its theory elegancy, strong framework compatibility and efficient manifold representations. 
The CVAEs commonly consist on an encoder that maps the input $\bm{x}$ and class $y$ to a pre-fixed distribution over the latent variable $\bm{z}$, and on a decoder that, given a latent variable $\bm{z}$ and the class $y$ tries to reconstruct the input $\bm{x}$. 
During training the model is trained by minimizing the negative variational lower bound of the conditional density of the data, defined as follows:
\begin{equation*}
    \begin{split}
    \mathcal{L} \big[ \bm{x}, y ; \theta, \phi \big] = \ &D_{\textrm{KL}} \Big[ q_{\phi} (\bm{z}|\bm{x})\|p(\bm{z}| y) \Big] \\
    &- \mathds{E}_{q_{\phi}(\bm{z}|x)} \Big[ \log p_\theta (\bm{x}|\bm{z}, y) \Big]
    \end{split}
\end{equation*}
where $q_\phi (\bm{z}|\bm{x})$ denotes the posterior of the encoder, and $p_\theta(\bm{z}| y)$ indicates the prior distribution over the latent variable $\bm{z}$ conditioned on the class $y$. 
The first term in the loss function is a regularizer that enforces the approximate posterior distribution $q_\phi(\bm{x}|\bm{z})$ to be close to the prior distribution $p_\theta(\bm{z}|y)$, while the second term is the average reconstruction error of the chained encoding-decoding process. The original VAE \cite{vae}, that uses an unconditioned prior distribution, presumes that $p_\theta(\bm{z})$ is an isotropic multivariate Gaussian $\mathcal{N}(\bm{0}, \bm{I})$ and $q_\phi(\bm{z}|\bm{x})$ is a general multivariate Gaussian $\mathcal{N}(\bm{\mu}, \bm{\sigma^2})$. With these assumptions, the KL-divergence term given a $K$-dimensional $\bm{z}$, can be computed in closed form and expressed as:
\begin{equation*}
    D_{\textrm{KL}}\Big[ q_\phi(\bm{z}|\bm{x})\|p_\theta(\bm{z}) \Big] = -\frac{1}{2}\sum_i^K (1+\log (\bm{\sigma}_i^2)-\bm{\mu}_i^2-\bm{\sigma}_i^2).
\end{equation*}
For the CVAE the KL-divergence term can be computed or estimated using only tractable latent prior distributions $p(\bm{z}|y)$~\cite{Pagnoni2018ConditionalVA}.
\subsection{CapsNet Formulation}

The capsule network, proposed by Hinton et al. \cite{tcap}, is a shallow architecture composed by two convolutional layers and two capsule layers. The first convolutional operation converts the pixel intensities of the input image $\bm{x}$ to primary local feature maps, while the second convolutional layer produces the primary capsules $\mathbf{u}_i$. Each capsule corresponds to a set of matrices that rotate the primary capsules for predicting the pose transformation $\hat{\mathbf{u}}_{j|i}=\mathbf{W}_{ij}\mathbf{u}_i$. Afterwards, the digit capsules $\mathbf{v}_j$, used for classification, are produced as the weighted sum of the primary capsules $\mathbf{v}_j=\sum_ic_{ij}\hat{\mathbf{u}}_{j|i}$, where the coefficients $c_{ij}$ are determined by the dynamic routing algorithm (DR), in which the primary capsules are compared to the digit capsules. For $t$-th iteration of DR, the coefficients are updated by,
\begin{equation*}
    \mathbf{c}^{(t+1)}_{i} = \mathbf{Softmax}(\mathbf{b}^{(t+1)}_i), ~~b^{(t+1)}_{ij} = b^{(t)}_{ij}+\hat{\mathbf{u}}_{j|i}\cdot \mathbf{v}^{(t)}_j.
\end{equation*}
For all layers of capsules, a squash function is used to introduce non-linearity and shrunk the length of capsule vectors into $[0,1]$,
\begin{equation*}
    \mathbf{Squash}(\mathbf{v})=\frac{\|\mathbf{v}\|^2}{1+\|\mathbf{v}\|^2}\frac{\mathbf{v}}{\|\mathbf{v}\|}.
\end{equation*}
In this way the norm of the capsule stands for the probability of a particular feature being present in the input image $\bm{x}$.

%% file: sections/s4_proposed_method_iccv.tex
\section{Proposed Method}


\begin{figure*}[!ht]
    \centering
    \includegraphics[width=1.\textwidth]{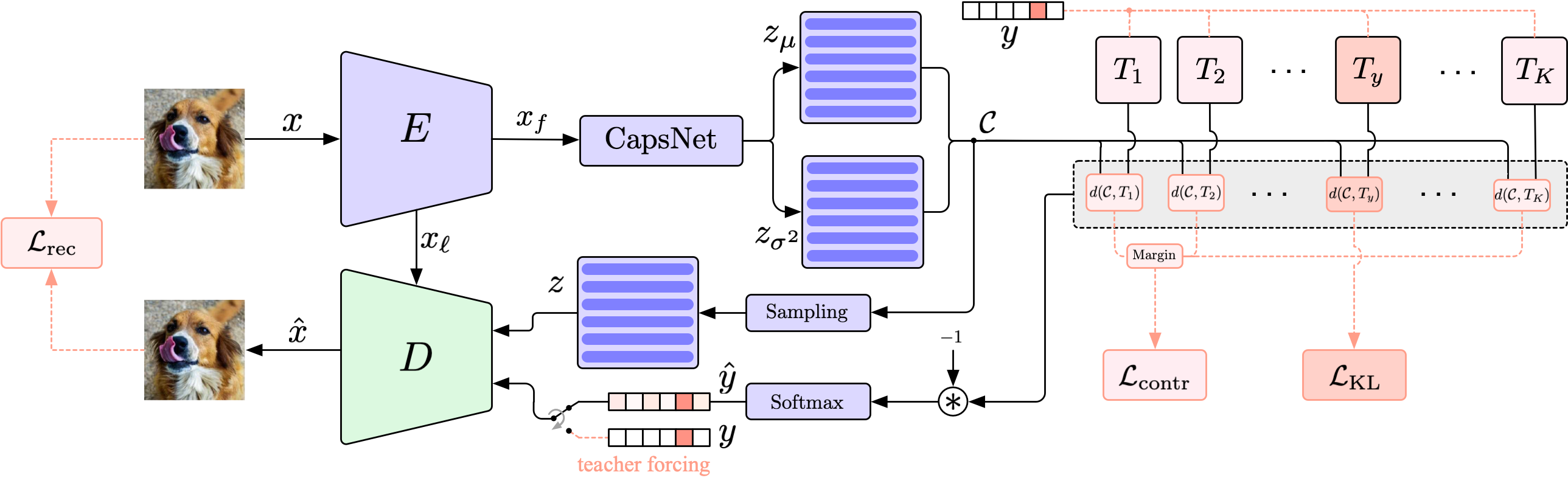}
    \caption{Outline of our CVAECapOSR model. \label{fig:architecture} The dashed orange lines stand for the computation of the model during training, whereas the solid black lines indicate the computation done by the model both during training and testing time.}
\end{figure*}

\subsection{Model Architecture}

Our model, depicted in Figure~\ref{fig:architecture}, is based on a CVAE with $K$ different gaussian prior distributions, one for each known class. Given the input image $\bm{x}$ and its corresponding label $y$ the encoder processes $\bm{x}$ producing the feature representation $\bm{x}_f$. Afterward the capsule network computes the distribution $q(\bm{z}|\bm{x})$ that is pushed toward the conditioned prior $p(\bm{z}|y) = \bm{T}_y$ during the learning process. Using the distance information between $q(\bm{z}|\bm{x})$ and all the targets we estimate the class $\hat{y}$, and using the reparametrization trick we sample $\bm{z}$ from $q(\bm{z}|\bm{x})$. Given $\hat{y}$ and $\bm{z}$ we compute the reconstruction $\hat{\bm{x}}$ through the decoder that is a convolutional neural network that uses transposed convolutions. After this general description of the computation of our model, we now present in deep the architecture step by step.

\textbf{Encoding Stage.} The blocks involved in the encoding stage are an encoder and a capsule network. The encoder is a convolutional neural network that processes the input image $\bm{x} \in \mathbb{R}^{C \times  H \times  W}$ producing the feature $\bm{x}_f \in \mathbb{R}^{d_c \times d_h \times  d_w}$. Then, similar to~\cite{sabour2017dynamic}, the capsule network processes the feature representation $\bm{x}_f$ by computing the primary capsules $\bm{x}_{\textrm{pc}} \in \mathbb{R}^{f_1 \times (\frac{d_c}{f_1} d_h d_w)}$ and then the digit capsules $\bm{x}_{\textrm{dc}} \in \mathbb{R}^{K \times f_2}$ using the dynamic routing algorithm. We indicate with $f_1$ the dimension of the primary capsules and with $f_2$ the dimension of the digit capsules. Given the capsules $\bm{x}_{\textrm{dc}}$ we compute the mean $\bm{z}_{\mu} \in \mathbb{R}^{K \times d}$ and variance $\bm{z}_{\sigma^2} \in \mathbb{R}^{K \times d}$ of the capsules distribution $\bm{\mathcal{C}} = q(\bm{z}|\bm{x})$ by applying a capsule-wise fully connected layer with $d$ output units. In this way the probabilistic capsule network produces $K$ mean capsules $\{ \bm{z}_{\mu}^{(k)} \}_{k=1}^K$ and $K$ variance capsules $\{ \bm{z}_{\sigma^2}^{(k)} \}_{k=1}^K$, each of size $d$.

\textbf{Contrastive Variational Stage.} We design each CVAE prior target $p(\bm{z}|y=k)$ to be a gaussian distribution $\bm{T}_k = \mathcal{N} \left( \bm{\tilde{\mu}}_k, \bm{\tilde{\Sigma}}_k \right)$ with learnable mean vector $\bm{\tilde{\mu}}_k \in \mathbb{R}^{Kd}$ and learnable diagonal covariance matrix $\bm{\tilde{\Sigma}}_k \in \mathbb{R}^{Kd \times Kd}$ with \mbox{$1 \leq k \leq K$}. In order to simplify the notation of our model framework we consider the targets $\bm{T}_k$ as gaussian distributions defined by $\bm{\mu}_k \in \mathbb{R}^{K \times d}$ being the reshaped version of the $\tilde{\bm{\mu}}_k$ and $\bm{\Sigma}_k \in \mathbb{R}^{K \times d}$ being the reshaped diagonal of $\tilde{\bm{\Sigma}}_k$. In order to map input instances from the same class into compacted and separated regions of the latent space, during the learning process, we let the probabilistic capsule $\bm{\mathcal{C}}$ to be attracted by the $y$-th target distribution $\bm{T}_y$ and at the same time we let all the other targets $\bm{T}_{\neq y}$ to be repulsed by $\bm{\mathcal{C}}$. Using this contrastive strategy, we encourage the encoded representation to belong to the correct region of the latent space while maintaining all the targets sufficiently far apart to each other. We then estimate the class $\hat{y}$ of the input sample $\bm{x}$ as:
\begin{equation*}
    \hat{y}_k = p(y = k| \bm{x}) = \frac{e^{- \gamma d(\bm{\mathcal{C}}, \bm{T}_k)}}{\sum_{j=1}^K e^{- \gamma d(\bm{\mathcal{C}}, \bm{T}_j)}}, \ \ \ 1\leq k\leq K,
\end{equation*} where 
\begin{equation*}
    d(\bm{\mathcal{C}}, \bm{T}_i) = \frac{1}{K}\sum_{k=1}^K D_{\textrm{KL}} \Big[ \bm{\mathcal{C}}^{(k)} || \bm{T}_i^{(k)} \Big],
\end{equation*}
is the distance between the probabilistic capsules $\bm{\mathcal{C}}$ and the target $\bm{T}_i$, and $\gamma$ is a coefficient parameter that controls the hardness of the probability assignment. In this way we estimate the probability of $\bm{x}$ being of class $k$ considering the whole configuration of capsules $\{ \bm{\mathcal{C}}^{(k)} \}_{k=1}^{K}$ and not only the single most activated capsule as done in~\cite{sabour2017dynamic}.

\textbf{Decoding Stage.} Given the class estimate $\hat{y} \in \mathbb{R}^{K}$ we compute its learnable embedding $\hat{y}_e = \textrm{Embedding}(\argmax_{k}(\hat{y}_k)) \in \mathbb{R}^{d}$, and given the sampled latent capsules $\bm{z} \in \mathbb{R}^{K \times d}$ we compute the reconstruction $\hat{\bm{x}}$ through the decoder starting from $\bm{z}_y \in \mathbb{R}^{K \times d}$ with $\bm{z}^{(k)}_y = \hat{y}_e + \bm{z}^{(k)}$. The decoder is a convolutional neural network with transposed convolutions that follows a symmetrical structure of the encoder. Similar to~\cite{Ronneberger2015UNetCN}, we implement lateral connections $\bm{x}_\ell$ with $1 \leq \ell \leq 4$ from the internal features of the encoder to the decoder that during training are randomly dropout for making the decoder less dependent from the internal representations of the encoder.

\subsection{Training}
We train the model on the closed dataset, and during the learning process for a single input sample $(\bm{x}, y)$ we minimize the following loss function:
\begin{equation}
    \mathcal{L}(\bm{x}, y) = \mathcal{L}_{\textrm{KL}} (\bm{x}, y) + \alpha \mathcal{L}_{\textrm{contr}}(\bm{x}, y) + \beta \mathcal{L}_{\textrm{rec}}(\bm{x}) ,
    \label{eq:loss}
\end{equation}
where
\begin{equation}
    \mathcal{L}_{\textrm{KL}} (\bm{x}, y) = d(\bm{\mathcal{C}}, \textrm{sg}\big[ \bm{T}_y \big]),
    \label{eq:kl}
\end{equation}
\begin{equation}
    \mathcal{L}_{\textrm{contr}} (\bm{x}, y) = \frac{1}{K-1}\sum_{k \neq y}^K \left[ m_k - d(\textrm{sg}\big[ \bm{\mathcal{C}} \big], \bm{T}_k) \right]^+
    \label{eq:contr},
\end{equation}
\begin{equation}
    \mathcal{L}_{\textrm{rec}} (\bm{x}) = \big\| \hat{\bm{x}} - \bm{x} \big\|_2^2.
    \label{eq:rec}
\end{equation}

As already defined in \cite{NIPS2017_7a98af17}, the function $\textrm{sg}[\cdot]$ in Eq.~\eqref{eq:kl} and Eq.~\eqref{eq:contr} stands for the stop-gradient operator that is defined as the identity at forward computation time and has zero partial derivatives, constraining its argument to be a non-updated constant. The loss term in Eq.~\eqref{eq:kl} is responsible for pushing the probabilistic capsules $\bm{\mathcal{C}}$ toward the target  $\bm{T}_y$, leading to the concentration of all the density of known samples in the targets region. On the other hand, the contrastive loss term in Eq.~\eqref{eq:contr} pushes all the targets not related to $y$ far away from the distribution $\bm{\mathcal{C}}$ using a margin loss with margin $m_k$ where $[\cdot]^+$ is the function that returns the positive part of its argument. By considering $\bm{T}_{\neq y}$ to be the otherness of $\bm{T}_y$, therefore of $\bm{\mathcal{C}}$, the contrastive term not only avoid the collapse of the prior targets, but also encourages the separation between one class and \textit{all the other} classes, potentially the unknown counterpart. Finally, the loss term in Eq.~\eqref{eq:rec} is the mean squared error reconstruction between the input and the output of the model. We aggregate all loss terms in Eq.~\eqref{eq:loss} and we control the strength of $\mathcal{L}_{\textrm{contr}}$ with a parameter $\alpha$ and the strength of $\mathcal{L}_{\textrm{rec}}$ with a parameter $\beta$. During training we found beneficial to use teacher forcing in the decoder i.e. we decided to feed $y$ instead of the estimate $\hat{y}$ to the decoder, while during validation and testing we feed only the estimated quantity, enabling in this way, the independence of our model respect to the label $y$ during inference.

\subsection{Inference}
We use the model’s natural rejection rule based on the probabilistic distance between samples and targets to detect unknowns, and directly classify known samples with the minimum probabilistic distance over than a given threshold.

Given a new sample $\bm{x}$ we decide if it is an outlier as follows:
\begin{equation*}
  \hat{y} = \left \{
  \begin{aligned}
    &K + 1, \ \ \text{if}\ \max_{k} \Big\{ d(\bm{\mathcal{C}}, \bm{T}_k) \Big\} = d^* < \tau &&\\
    &\argmax_{k} \Big\{ d(\bm{\mathcal{C}}, \bm{T}_k) \Big\}, \ \ \text{otherwise}. &&
  \end{aligned} \right.
\end{equation*} 
where $\tau$ is found using cross validation, and $K+1$ is the new, unknown class not seen during training.

%% file: sections/s5_experimental_results_iccv.tex
\section{Experiments}

Recent works in this area followed the protocol presented in~\cite{Neal_2018_ECCV}. In that work, an open set recognition scenario is obtained by randomly selecting $K$ classes from a specific dataset as known (see below for more details), while the remaining classes are considered to be open set classes. This procedure is applied to five random splits.
However, as recently shown by~\cite{Perera_2020_CVPR}, performance across different splits varies significantly (e.g. AUROC on CIFAR10 varied between $77\%$ to $87\%$ across different splits), and there are serious reproducibility issues.
Moreover, not only the splits have a large influence on the results, but also the strategy used to select the samples belonging to the unknown classes.
Therefore, starting from the splits used in~\cite{Neal_2018_ECCV} and following~\cite{Perera_2020_CVPR}, we publicly release our code and data\footnote{Code and data publicly available on \url{https://github.com/guglielmocamporese/cvaecaposr}.}, as well as the implementation of other state-of-the-art methods, to foster a fair comparison on this task.


\begin{table*}[th]
    \centering
    \begin{tabular}{l|c|c|c|c|c|c} \hline
    Method     &   MNIST & SVHN & CIFAR10 & CIFAR+10 & CIFAR+50 & TinyImageNet \\
    \hline
    Softmax~\dag~\cite{cgdl} & 0.978 & 0.886 & 0.677 & 0.816 & 0.805 & 0.577 \\
    Openmax~\dag~\cite{topn} & 0.981 & 0.894 & 0.695 & 0.817 & 0.796 & 0.576 \\
    G-Openmax~\dag~\cite{gopenmax} & 0.984 & 0.896 & 0.675 & 0.827 & 0.819 & 0.580 \\
    OSRCI~\dag~\cite{Neal_2018_ECCV} & 0.988 $\scriptstyle{\pm 0.004}$ & 0.91 $\scriptstyle{\pm 0.01}$ & 0.699 $\scriptstyle{\pm 0.038}$ & 0.838 & 0.827 & 0.586 \\
    CROSR~\cite{crosr}    & 0.991 $\scriptstyle{\pm 0.004}$ &  0.899 $\scriptstyle{\pm  0.018}$ & - & - & - & 0.589 \\
    C2AE~\ddag~\cite{c2ae}       & - & 0.892 $\scriptstyle{\pm 0.013}$ & 0.711 $\scriptstyle{\pm 0.008}$ & 0.810 $\scriptstyle{\pm 0.005}$ & 0.803 $\scriptstyle{\pm 0.000}$ & 0.581 $\scriptstyle{\pm 0.019}$ \\
    GFROR~\ddag~\cite{Perera_2020_CVPR}  & - & 0.955 $\scriptstyle{\pm 0.018}$ & 0.831 $\scriptstyle{\pm 0.039}$ & - & - & 0.657 $\scriptstyle{\pm 0.012}$ \\
    CGDL \S~\cite{cgdl} & 0.977 $\scriptstyle{\pm 0.008}$ & 0.896 $\scriptstyle{\pm 0.023}$ & 0.681 $\scriptstyle{\pm 0.029}$ & 0.794 $\scriptstyle{\pm 0.013}$ & 0.794 $\scriptstyle{\pm 0.003}$ & 0.653 $\scriptstyle{\pm 0.002}$ \\ 
    RPL \S~\cite{rpl}  & 0.917 $\scriptstyle{\pm 0.006}$ & 0.931 $\scriptstyle{\pm 0.014}$ & 0.784 $\scriptstyle{\pm 0.025}$& 0.885 $\scriptstyle{\pm 0.019}$ & 0.881 $\scriptstyle{\pm 0.014}$ & 0.711 $\scriptstyle{\pm 0.026}$ \\
    \hline
    CVAECapOSR (ours)       & \textbf{0.992} $\scriptstyle{\pm 0.004}$ & \textbf{0.956} $\scriptstyle{\pm 0.012}$ & \textbf{0.835} $\scriptstyle{\pm 0.023}$ & \textbf{0.888} $\scriptstyle{\pm 0.019}$ & \textbf{0.889} $\scriptstyle{\pm 0.017}$ &  \textbf{0.715}  $\scriptstyle{\pm 0.018}$ \\ \hline
    \end{tabular}
    \caption{AUROC scores on the detection of known and unknown samples. Results are averaged over $5$ different splits of known and unknown classes partitions. As discussed in Section~\ref{sec:results}, we report the results on the same data splits and, for the sake of clarity, we highlight the source of the results used to populate the table: $\dag$ are provided by~\cite{Neal_2018_ECCV}, $\ddag$ is from~\cite{Perera_2020_CVPR} and $\S$ are the results that we obtained by running the code of the original paper.}
    \label{tab:auc}
\end{table*}

\begin{table*}[th]
    \centering
    \begin{adjustbox}{width=\linewidth,center}
	\setlength{\tabcolsep}{5pt}
    \begin{tabular}{l|c|c|c|c|c|c} \hline
    Method     &   MNIST & SVHN & CIFAR10 & CIFAR+10 & CIFAR+50 & TinyImageNet \\
    \hline
    CVAECapOSR fixed Targets & \textbf{0.997} $\scriptstyle{\pm 0.006}$ & 0.953 $\scriptstyle{\pm 0.022}$ & 0.823 $\pm$ $\scriptstyle{\pm 0.012}$ & 0.868 $\scriptstyle{\pm 0.018}$ & 0.829 $\scriptstyle{\pm 0.009}$ &  0.706  $\scriptstyle{\pm 0.014}$ \\ \hline
    CVAECapOSR learn Targets  & 0.992 $\scriptstyle{\pm 0.004}$ & \textbf{0.956} $\scriptstyle{\pm 0.012}$ & \textbf{0.835} $\pm$ $\scriptstyle{\pm 0.023}$ & \textbf{0.888} $\scriptstyle{\pm 0.019}$ & \textbf{0.889} $\scriptstyle{\pm 0.017}$ &  \textbf{0.715}  $\scriptstyle{\pm 0.018}$ \\ \hline
    \end{tabular}
    \end{adjustbox}
    \caption{AUROC scores on the detection of known and unknown samples comparing our model that uses fixed targets (first row) versus  our model that learns the targets (second row) during the learning process. Results are averaged over $5$ splits.}
    \label{tab:auc_ablation}
\end{table*}

\subsection{Datasets}
We evaluate open set recognition performance on the standard datasets used in previous works, i.e. MNIST~\cite{lecun-mnisthandwrittendigit-2010}, SVHN~\cite{Netzer2011ReadingDI}, CIFAR10~\cite{Krizhevsky2009LearningML}, CIFAR+10, CIFAR+50 and TinyImageNet~\cite{Le2015TinyIV}.

\textbf{MNIST, SVHN, CIFAR10.} All three datasets contain ten categories. MNIST consists of hand-written digit images, and it has $60,000$ $28\times28$ grayscale images for training and $10,000$ for testing.
SVHN contains street view house numbers, consisting of ten digit classes each with between $9,981$ and $11,379$ $32\times32$ color images.
Then we consider the CIFAR10 dataset, which has $50,000$ $32\times32$ color images for training and $10,000$ for testing.
Following~\cite{Neal_2018_ECCV}, in the unknown detection task each dataset is randomly partitioned into $6$ known classes and $4$ unknown classes. In this setting, the openness score is fixed to $22.54\%$.

\textbf{CIFAR+10, CIFAR+50.} To test our model in a setting of higher openness values, we perform CIFAR+$Q$ experiments using CIFAR10 and CIFAR100~\cite{Krizhevsky2009LearningML}. To this end, $4$ known classes are sampled from CIFAR10 and $Q$ unknown classes are drawn randomly from the more diverse and larger CIFAR100 dataset. Openness scores of CIFAR+10 and CIFAR+50 are $46.54\%$ and $72.78\%$, respectively.

\textbf{TinyImageNet.} For TinyImagenet dataset, which is a subset of ImageNet that contains $200$ classes, we randomly sampled $20$ classes as known and the remaining classes as unknown. In this setting, the openness score is $68.37\%$.

\subsection{Metrics}

Open set classification performance is usually measured using F-score and AUROC (Area Under ROC Curve)~\cite{Geng2020RecentAI}. 
F-score is used to measure the in-distribution classification performance, while AUROC is commonly reported by both open set recognition and out-of-distribution detection literature.
AUROC provides a calibration free measure and characterizes the performance for a given score by varying the discrimination threshold \cite{prroc}.
In our experiments, we use macro averaged F1-score on the open set recognition task, and the AUROC for the unknown detection task.
For both metrics, higher values are better.

\subsection{Experimental Results}
\label{sec:results}

Following \cite{cgdl}, we conducted two major experiments in which our model has to solve the \textit{unknown detection} task and the \textit{open set recognition} task. For all the experiments we use ResNet34~\cite{resnet} as the encoder backbone of our model.

\textbf{Unknown Detection.} In the unknown detection problem the model is trained on a subset of the dataset using $K$ classes, and the evaluation is done by measuring the model capability on detecting unknown classes, not seen during training. The evaluation is performed by considering the binary recognition task of the known \emph{vs} unknown classes, and performances are reported in terms of AUROC scores. The results, shown in Table~\ref{tab:auc}, are averaged over five random splits of known and unknown classes, provided by~\cite{Neal_2018_ECCV}.

As already discussed (and recently shown in~\cite{Geng2020RecentAI,Perera_2020_CVPR}), performance across different splits varies significantly.
For this reason we use the exact data splits provided by~\cite{Neal_2018_ECCV} that have been used also in other recent works~\cite{crosr,Perera_2020_CVPR}.
Nevertheless, not all the results reported in these works are directly comparable; although the splits are the same, \cite{Perera_2020_CVPR} followed a particular strategy in selecting the open set classes in CIFAR+10 and CIFAR+50 experiments (i.e. they selected 10 and 50 samples from vehicle classes instead of purely random classes, which has gained a large impact on these results).
Therefore, following~\cite{Perera_2020_CVPR}, we have run the code of~\cite{cgdl} and~\cite{rpl} (whereas the results of~\cite{c2ae} are provided by~\cite{Perera_2020_CVPR}, since the code is no more available), and we compare all the results with the state of the art papers that have the same splits and use the same evaluation setting, and that can be reproduced.
As shown in Table~\ref{tab:auc}, we obtain state of the art results, outperforming all previous methods, on all the datasets.
Moreover, as also previously reported, we will release all data and code to guarantee reproducibility.

One important fact we observed during the training process is the boost we obtained by letting the targets distributions $\bm{T}_k$ to be learned and not to be used as fixed priors, as shown in Table~\ref{tab:auc_ablation}. We initialized the learnable targets with $\bm{\mu}_k^{(i)} = \mathbb{1}_{d} \cdot \delta_{k=i}$ and $\bm{\Sigma}_k^{(i)} = \mathbb{1}_d$. We noticed that learning the targets without considering the contrastive term in the loss function ($\alpha=0$) caused the collapse of the targets into one single distribution, leading to poor results. We thus consider the contrastive term, and we set $\alpha = 1.0$, $\beta = 0.05$ and $m_k = 10$.


\textbf{Open Set Recognition.} In the open set recognition problem, the model is trained on the closed dataset that contains $K$ classes, and it is evaluated on the open dataset considering $K + 1$ classes.
In this experimental setting, we evaluate the model using the macro F1-score on the $K + 1$ classes. In the first experiment for open set recognition, we train on all the classes of the MNIST dataset and we then evaluate the performances by including new datasets in the open set.
Similarly to~\cite{crosr}, we used Omniglot, MNIST-Noise, and Noise that are datasets of gray-scale images. Each of this dataset contains $10,000$ test images, the same as MNIST. 
The Omniglot dataset contains hand-written characters from the alphabets of many languages, while the Noise dataset has images synthesized by randomly sampling each pixel value independently from a uniform distribution on $[0, 1]$. MNIST-Noise is also a synthesized set, constructed by superimposing MNIST’s test images on Noise. The results of the open set recognition on these datasets are shown in Table~\ref{mnist}. On each dataset, we outperform state of the art results by a large margin. On Omniglot, we improve the F1-score by $+0.121$, in the MNIST-Noise by $+0.095$, and in the Noise by $+0.123$.

\begin{table}[t]
    \centering
    \begin{adjustbox}{width=\linewidth,center}
    \setlength{\tabcolsep}{5pt}
    \begin{tabular}{l|c|c|c}\hline
     Method  &  Omniglot & MNIST-noise & Noise  \\ \hline
     Softmax~\cite{cgdl} &  0.595    & 0.801       & 0.829  \\
     Openmax~\cite{gopenmax} &  0.780    & 0.816       & 0.826  \\
     CROSR~\cite{crosr}   &  0.793    & 0.827       & 0.826  \\
     CGDL~\cite{cgdl}    &  0.850    & 0.887       & 0.859  \\ \hline
     CVAECapOSR (ours)    &  \textbf{0.971}   & \textbf{0.982}      & \textbf{0.982} \\ \hline
    \end{tabular}
    \end{adjustbox}
    \caption{Results for the open set recognition on the MNIST dataset. We report the macro-averaged F1-score for $11$ classes ($10$ from the test partition of the MNIST, and $1$ from the test of another dataset).}
    \label{mnist}
\end{table}

\begin{table}[t]
    \centering
    \begin{adjustbox}{width=\columnwidth,center}
	\setlength{\tabcolsep}{2pt}
    \begin{tabular}{l c c c c}\hline
    \multicolumn{5}{ c }{AUROC scores w.r.t. different Openness Values ($O$)}\\
    \hline
    \hline
     Openness Variation:   &$O=0\%$     &$O=15.98\%$   &$O=30.72\%$   &$O=39.69\%$ \\
     \hline
     \hline
     \multicolumn{5}{ c }{Influence of the Feature Extractor (before CapsNet)}\\
     \hline
     CapsNet &0.971 &0.753 &0.767 &0.781 \\
     ResNet20 + CapsNet     &\textbf{0.981}   &\textbf{0.948}     &\textbf{0.949}     &\textbf{0.950}   \\
     \hline
     Improvement     &+0.020   &+0.195	  &+0.182	 &+0.169   \\
     \hline
     \hline
     \multicolumn{5}{ c }{Influence of CapsNet}\\
     \hline
     ResNet20 + FC            &0.975   &0.581	 &0.595	   &0.606   \\
     ResNet20 + CapsNet     &\textbf{0.981}   &\textbf{0.948}          &\textbf{0.949}&\textbf{0.950}   \\
     \hline
     Improvement     &+0.006   &+0.367	  &+0.354	    &+0.344   \\
     \hline
     \hline
     \multicolumn{5}{ c }{Influence of Dynamic Routing}\\
     \hline
     ResNet20 + CapsNet     &0.981   &0.948     &0.949     &0.950   \\
     ResNet20 + CapsNet + DR  &\textbf{0.982}    &\textbf{0.952}       &\textbf{0.954}   &\textbf{0.955}\\ 
     \hline
     Improvement     &+0.001   &+0.004	  &+0.005	    &+0.005   \\
     \hline
    \end{tabular}
    \end{adjustbox}
    \caption{Ablation study on the model architecture. We report results for the unknown detection task on SVHN dataset with outliers from CIFAR100. The performance is evaluated by AUROC with different openness values.}
    \label{tab:abalation}
\end{table}

\begin{figure*}[!htp]
    \centering
    \begin{tabular}{c c c}
    
        \begin{subfigure}[b]{0.39\textwidth}
            \centering
            \caption{CVAECapOSR with Capsules}
            \includegraphics[width=\textwidth]{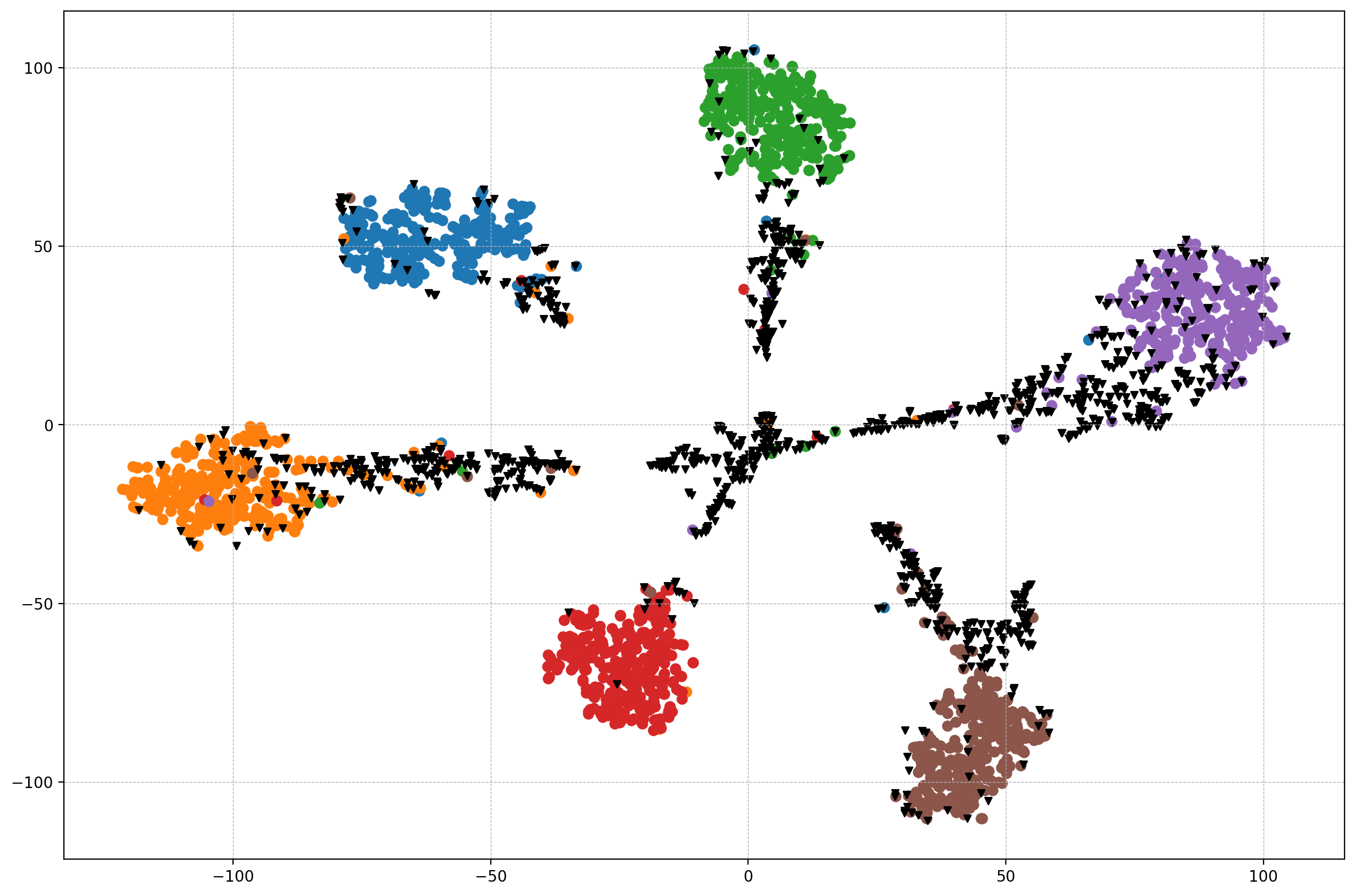}
            
            \label{fig:t-sne_omniglot}
         \end{subfigure} &

         \begin{subfigure}[b]{0.39\textwidth}
            \centering
            \caption{CVAECapOSR with Standard Neurons}
            \includegraphics[width=\textwidth]{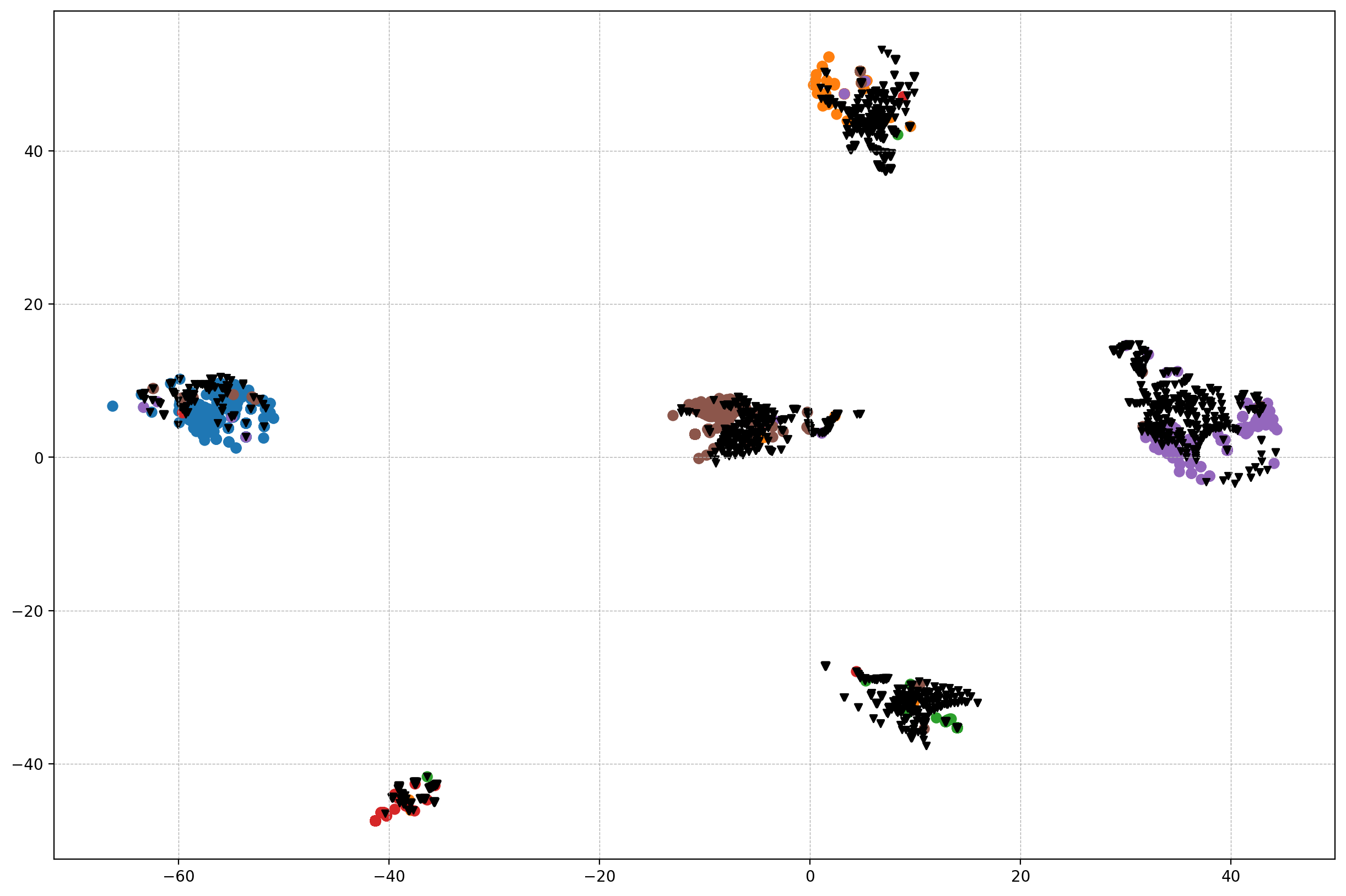}
            
            \label{fig:t-sne_mnist_noiet}
         \end{subfigure} &

        \begin{subfigure}[t]{0.14\textwidth}
            \centering
            \includegraphics[width=\textwidth]{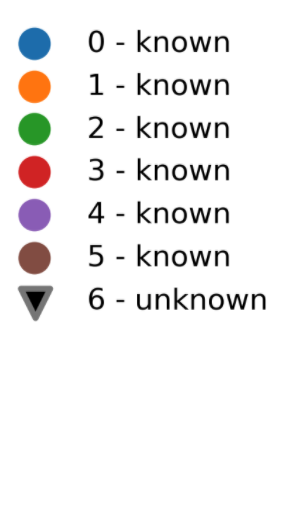}
         \end{subfigure} \\

    \end{tabular}
    \caption{t-SNE latent space visualizations obtained with different components (i.e., (a) Capsules, (b) Standard Neurons) of the SVHN test set including the CIFAR+100 test set from which unknown samples are sampled. In particular, we use an openness value of $O = 35.67\%$. In both pictures, the unknown samples are represented by black triangles.}
    \label{fig:t-sne_ablation_svhn}
\end{figure*}

In the second experiment of open set recognition, following the same protocol used in~\cite{Neal_2018_ECCV}, all samples from the $10$ classes in CIFAR10 dataset are considered as known data, and samples from ImageNet and LSUN are selected as unknown samples.
In order to have the same image size as known samples, we resized or croped the unknown samples, obtaining the following datasets: ImageNet-crop, ImageNet-resize, LSUN-crop, and LSUN-resize. For each dataset, we consider all their $10,000$ test samples as the unknown samples in the open set. The performance of the method is evaluated using the macro-averaged F1-scores in the $11$ classes ($10$ known classes and 1 unknown), and the results are shown in Table~\ref{tab:cifar}. We can see that our method outperforms all previous methods under the F1-score on ImageNet-resize, ImageNet-crop, LSUN-crop, and LSUN-resize.

\begin{table*}
    \centering
    \begin{adjustbox}{width=0.9\linewidth,center}
	\setlength{\tabcolsep}{10pt}
    \begin{tabular}{l|c|c|c|c}\hline
    Method             & ImageNet-crop & ImageNet-resize & LSUN-crop & LSUN-resize \\
    \hline
    Softmax $\dag$~\cite{cgdl}            & 0.639         &0.653  &0.642  &0.647 \\
    Openmax $\dag$~\cite{topn}            & 0.660         &0.684  &0.657  &0.668 \\
    CROSR~\cite{crosr}             & 0.721         &0.735  &0.720  &0.749 \\
    C2AE $\ddag$~\cite{c2ae}               & 0.837         &0.826  &0.783  &0.801 \\
    CGDL $\S$~\cite{cgdl}               & 0.840         & 0.832  & 0.806 & 0.812 \\
    RPL $\S$~\cite{rpl}               & 0.811         & 0.810  & 0.846 & 0.820 \\
    \hline
    CVAECapOSR (ours) & \textbf{0.857} & \textbf{0.834}  & \textbf{0.868} & \textbf{0.882} \\
    \hline
    \end{tabular}
    \end{adjustbox}
    \caption{Open set recognition results on CIFAR-10 with various outliers added to the test set as unknowns. We evaluate the model using macro-averaged F1-scores on $11$ classes ($10$ from the the test of the CIFAR10, and $1$ from various test datasets). For the sake of clarity, we highlight the source of the results used to populate the table: $\dag$ are provided by~\cite{Neal_2018_ECCV}, $\ddag$ is from~\cite{Perera_2020_CVPR} and $\S$ are the results that we obtained by running the code of the original paper }
    \label{tab:cifar}
\end{table*}

\textbf{Ablation Study on the Model Architecture.} In order to verify the contribution of each part of our model, we perform ablation on the relevance of the main model's components: the capsule network CapsNet, and the feature extractor. ResNet20 is selected as the feature extractor for getting shorter training times. We investigate also the impact of different components of CapsNet, in order to understand their importance. We consider four different variations of our model architecture: the model with CapsNet and dynamic routing that does not use the ResNet20 feature extractor, but just a single convolutional layer; ResNet20+CapsNet that includes the residual feature extractor before CapsNet and doesn't use dynamic routing; ResNet20+FC where a fully connected layer replaces CapsNet; and ResNet20+CapsNet+DR that implements dynamic routing in CapsNet. For all CapsNets that do not implement the dynamic routing, we process each capsule by a fully connected layer. For the ablation, we consider the entire SVHN dataset as the closed dataset, and we consider unknown samples from the CIFAR100 for the open dataset. We then consider different numbers of unknown classes of CIFAR100, leading to different openness values $O$. The results of the ablation analysis are reported in Table~\ref{tab:abalation}. We can see that the residual network used as a feature extractor in the encoder helps, especially when the openness $O$ of the open set increases. This fact highlights the importance of having already pre-processed features for the CapsNet on the unknown detection problem when openness increases. Furthermore, another emerging fact is the higher representation capability of the CapsNet with respect to a FC layer: as the openness increases, the AUROC improvement increases up to $+0.344$. This result suggests that capsules are more capable in detecting unknown samples with respect to standard artificial neurons. This fact emerges also from the t-SNE~\cite{Maaten2008VisualizingDU} visualization of the latent space, reported in Figure~\ref{fig:t-sne_ablation_svhn}, where the separation between known and unknown produced by the probabilistic capsules is more evident with respect to the one produced by a standard FC. 
Finally, from the experiments we see that dynamic routing achieves better performances with respect to not using it, and that the largest boost on using the capsule network is given by the pose transformation.

\textbf{Implementation details.} We also investigated the importance of the parameters $\alpha$, $\beta$, $m_k$ and $\gamma$. To this end, we conducted the unknown detection experiment with the ResNet20+CapsNet+DR architecture on SVHN, using CIFAR100 as the open dataset with openness $\mathcal{O}=30.72\%$. As suggested by the results reported in Table~\ref{alpha_mk}, we set $\alpha = 1.0$ and $m_k = 10.0$ and, finally, we set $\beta=0.05$ and $\gamma=1$ empirically.

\begin{table}[t]
    \centering
    \begin{adjustbox}{width=0.9\linewidth,center}
    \setlength{\tabcolsep}{3pt}
    \begin{tabular}{c|c|c|c}\hline
     Contr. Params  &  $m_k=5.0$ & $m_k=10.0$ & $m_k=20.0$\\ \hline
     $\alpha=0.5$ & \cellcolor{red_table}0.527 & \cellcolor{red_table}0.564 & \cellcolor{green_table}0.947  \\
     $\alpha=1.0$ & \cellcolor{green_table}0.937 & \cellcolor{green_table}0.954 & \cellcolor{green_table}0.949  \\
     $\alpha=2.0$ & \cellcolor{green_table}0.944 & \cellcolor{green_table}0.951 & \cellcolor{green_table}0.945  \\
    \hline
    \end{tabular}
    \end{adjustbox}
    \caption{AUROC scores for different values of the parameters $\alpha$, $m_k$ in the loss function. Red cells indicate that targets during the learning process overlap at some point, leading to poor results. Green cells indicate no collapse of prior targets, suggesting good values for $\alpha$, and $m_k$.}
    \label{alpha_mk}
\end{table}

%% file: sections/s6_conclusion_iccv.tex
\section{Conclusion}
In this paper, we introduced CVAECapOSR, a model for open set recognition based on CVAE that produces probabilistic capsules as latent representations through the capsule network. We extended the standard framework of CVAEs using multiple gaussian prior distributions rather than just one for all known classes in the closed dataset. Furthermore, targets are set to be learnable in order to cluster knowns inside their target regions. The contrastive term is used to model the otherness for known classes and to keep the target regions to be mutually separated. 
Experimental results, obtained on several datasets, show the effectiveness and the high performances 
on unknown detection and open set recognition tasks.